\documentclass{article}

\usepackage[preprint,nonatbib]{neurips_2022}

\usepackage[utf8]{inputenc} 
\usepackage[T1]{fontenc}    
\usepackage[pagebackref=true,breaklinks=true,colorlinks,citecolor=citecolor, linkcolor=linkcolor]{hyperref}
\usepackage{url}            
\usepackage{booktabs}       
\usepackage{amsfonts}       
\usepackage{amsmath}
\usepackage{nicefrac}       
\usepackage{microtype}      
\usepackage[table]{xcolor}  
\usepackage{tabularx}
\usepackage{graphicx}
\usepackage{multirow}
\usepackage{enumitem}
\usepackage{pifont}
\usepackage{booktabs}
\usepackage{threeparttable}

\newcommand{\cmark}{\ding{51}}%
\newcommand{\xmark}{\ding{55}}%
\newcommand{\ccmark}{\ding{170}}%
\newcommand{\cccmark}{\ding{171}}%

\def\myspace{\hskip0.75em\relax}

\definecolor{citecolor}{HTML}{0071BC}
\definecolor{linkcolor}{HTML}{ED1C24}
\usepackage[ruled,vlined]{algorithm2e}

\newtheorem{definition}{Definition}

\usepackage{graphicx}
\usepackage{subcaption}
\title{  Phoenix: Democratizing ChatGPT  across Languages } 
\usepackage{CJKutf8}

\author{  Zhihong Chen$^{1,2}$, Feng Jiang$^{1}$, Junying Chen$^{1,2}$, Tiannan Wang$^{1,2}$, Fei Yu$^{1}$,  Guiming Chen$^{1}$ \\ \textbf{Hongbo Zhang}$^{1,2}$ ,  \textbf{Juhao Liang}$^{1,2}$, \textbf{Chen Zhang}$^{1}$, \textbf{Zhiyi Zhang}$^{1}$,   \textbf{Jianquan Li}$^{1}$, \textbf{Xiang Wan}$^{1,2}$\\ \textbf{Benyou Wang}$^{1,2}$\thanks{Benyou is the corresponding author}~, \textbf{Haizhou Li}$^{1,2}$\\
$^1$ School of Data Science, The Chinese University of Hong Kong, Shenzhen \\
$^2$ Shenzhen Research Institute of Big Data \\
  \texttt{wangbenyou@cuhk.edu.cn} }

\begin{document}
\begin{CJK}{UTF8}{gkai}
\maketitle

\begin{figure}[h]
    \centering
     \vspace{-20pt}
    \includegraphics[width=4cm]{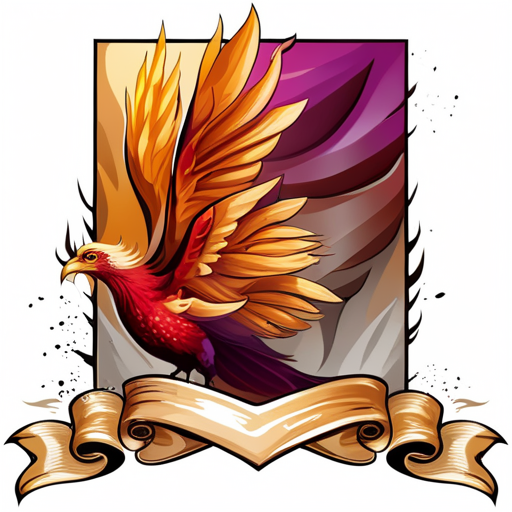}
\end{figure}

\begin{abstract}
This paper presents our efforts to democratize ChatGPT across language. We release a large language model `Phoenix', achieving competitive performance among open-source English and Chinese models while excelling in languages with limited resources (covering both Latin and non-Latin languages). We believe this work will be beneficial to make ChatGPT more accessible, especially in countries where people cannot use ChatGPT due to restrictions from OpenAI or local goverments. Our data, code, and models are available at \url{https://github.com/FreedomIntelligence/LLMZoo}.


\end{abstract}
\section{Introduction}
Nowadays, ChatGPT and its successor GPT-4 were developed and maintained by a single company, which unexpectedly results in `AI Supremacy' as defined below.
This is unacceptable for the AI community and may even lead to individual influence on the direction of the human future, thus bringing various hazards to society.

\begin{definition}[AI supremacy]
`AI supremacy' refers to a company's absolute leadership and monopoly position in an AI field, which may even include exclusive capabilities beyond general artificial intelligence. 
\end{definition}

As expressed in the widely-recognized Asilomar AI Principles, the development of advanced artificial intelligence has the potential to bring about a significant and transformative shift in the history of life on Earth.\footnote{\url{https://futureoflife.org/open-letter/pause-giant-ai-experiments/}} Therefore, the existence of AI supremacy could result in an unexpected consequence that the future of human beings (even all alive animals or plants) will be controlled by a single company; the responsibility of such a company might not be well-controlled.

\textbf{Make AI open again.}
Therefore, we aim to lower the cost and barrier of the ChatGPT training so that more responsible researchers can join the ChatGPT research and share their diverse thoughts, like figuring out \textit{how it works}, \textit{why it works}, and more importantly, \textit{how to develop large language models (like ChatGPT) in a planet-safe way}. This process is called democratization for the access and study of LLMs in~\cite{touvron2023llama}, where \cite{abacha2019medvqa,belle2023exploring,koala_blogpost_2023,vicuna2023} are among the process, see Sec.~\ref{sec:related_work} for more details.

\subsection{Methodology}
The existing open-source efforts~\cite{touvron2023llama} to democratize ChatGPT access explicitly exclude non-Latin and non-Cyrillic languages. This is definitely inconsistent with the open-source spirit. Imagine that one could decide not to allow a group of people to use light bulbs and vaccines -- most of us (even those who could use bulbs and vaccines) should be offended.

Therefore, this work is among the efforts to \textbf{democratize ChatGPT across Languages}. Currently, there are two lines of work to develop democratized ChatGPT\footnote{In the rest of our paper, ChatGPT does not refer to a specific product developed by OpenAI (called `OpenAI ChatGPT'), but a series of large language models designed for dialogue which might be from any companies. }.

\begin{itemize}
    \item (I) \textbf{Instuction-basd Tuning}. Instruction Tuning aims to \textit{tame} language models to follow human instructions \cite{ouyang2022training}, which might be manually designed, or in a hybrid fashion in that humans write some seed instructions and OpenAI ChatGPT is used to generate more similar instructions using in-context learning~\cite{wang2022self}.
    \item (II) \textbf{Conversation-basd Tuning}. ChatGPT-distilled conversations are used to teach language models to chat like OpenAI ChatGPT ~\cite{vicuna2023} while the instruction data is usually for single-turn question answering.
\end{itemize}

Existing models either lack open-source availability or are focused solely on English. There are very few models tailored to non-Latin languages, which makes it difficult for users in those languages to find suitable options.

\textbf{Phylosophy of methodology.}
We follow the two lines of work to train our \textit{multi-lingual} democratized ChatGPT. The key difference in our models is that we utilize two sets of data, namely \textit{instructions} and \textit{conversations}, which were previously only used by Alpaca and Vicuna, respectively. We believe incorporating both data types is essential for a recipe to achieve a proficient language model. The rationale is that the \textit{instruction} data helps to tame language models to adhere to human instructions and fulfill their information requirements, while the \textit{conversation} data facilitates the development of conversational skills in the model. These two types of data complement each other to create a more well-rounded language model. Another core idea is to leverage the multi-lingual nature inside the data, including the pre-training and instruction-tuning stage, where we start from the multi-lingual pre-trained backbone, BLOOM~\cite{workshop2023bloom}, and finetune it in the multi-lingual instruction-following data, without excluding any language.


\textbf{Training protocol.}
Particularly, the main challenge is to gather sufficient multi-lingual data of both types. To address this, for instruction-following data, we collect language-agnostic instructions and translate them into other languages according to the probability distribution of realistic languages using two ways, i.e., post-translation or post-answering. The former can confirm the answer quality, and the latter can introduce language-/culture-specific answers.\footnote{In general, the answers generated by ChatGPT in English are more robust, and the answers generated in other languages have more language-/culture-specific characteristics.} For conversation data, we collect user-shared conversations from various sources and in multiple languages. By such a design, we can train our model in a multi-lingual setting.

\textbf{Phylosophy to name `Phoenix'.}
The biggest barrier to developing LLMs is that we do not have enough candidate names for LLMs, as LLAMA, Guanaco, Vicuna, and Alpaca have already been used, and there are no more members in the camel family. 
We name our model `Phoenix '. In Chinese culture, the Phoenix is commonly regarded as a symbol of the king of birds; as the saying goes "百鸟朝凤", indicating its ability to coordinate with all birds, even if they speak different languages. We refer to Phoenix as the one capable of understanding and speaking hundreds of (bird) languages.\footnote{More importantly, Phoenix is the totem of ``the Chinese University of Hong Kong, Shenzhen'' (CUHKSZ).}

\textbf{A tailored `Phoenix' that is specific to the Latin language is called `Chimera'.}
Chimera is a similar hybrid creature in Greek mythology, composed of different Lycia and Asia Minor animal parts. Phoenix and Chimera are two legendary creatures standing for Eastern and Western cultures, respectively. We placed them in a zoo with the wish of great collaboration to democratize ChatGPT.

\subsection{Results}
\textbf{Evaluation Protocols.} We evaluate existing open-source models and `Phoenix' in automatic and manual ways. We collect 100 questions spanning ten categories and feed them to each model to get the answers. For automatic evaluation, we use GPT-4 as a reviewer to rate each answer from helpfulness, relevance, accuracy, and level of detail. For human evaluation, we ask the evaluation participants to judge the overall performance of the generation results of each model.

\textbf{The performance of Phoenix.} In Chinese, Phoenix achieves state-of-the-art performance among open-source large language models (e.g., BELLE and Chinese-LLaMA-Alpaca).\footnote{We exclude ChatGLM-6B since its training details and data are transparent. Therefore, it is impossible to replicate it from scratch. In this paper, we categorize ChatGLM-6B under non-open source models.} In other non-Latin languages, Phoenix largely outperforms existing LLMs in many languages, including Arabic, Japanese, and Korean.

\begin{definition}[multilingual tax]
    A multi-lingual model,  with a limited size, may not perform as well as a language-specific model when performing tasks specific to a particular language. This is because the multi-lingual model is designed to adapt to many languages, and some of its training may not be optimized for the specific language. As a result, language-specific models may be more accurate and efficient when dealing with tasks specific to a particular language.
\end{definition}

Among open-source Latin language models (e.g., Vicuna~\cite{vicuna2023}), Phoenix did not achieve state-of-the-art results owing to the fact that Phoenix additionally paid a `multi-lingual tax' when dealing with non-Latin or non-Cyrillic languages. As democratization itself cares about minor groups who speak relatively low-source languages, we believe such a `multi-lingual tax' for minor languages is worthy of paying. On the other hand, texts in various languages might share some commonness, so information and knowledge behind multi-lingual languages might be transferable. This gives multi-lingual LLMs additional merit to process cross-culture tasks in more comprehensive tasks. In some senses, this underscores the value of linguistic diversity and the need to consider the perspectives of individuals from diverse linguistic backgrounds, especially people who speak minor languages.

\textbf{Tax-free Phoenix: Chimera}. To reduce the multi-lingual tax in Latin and Cyrillic languages, we replace the backbone of Phoenix with LLaMA. In the English benchmark, Chimera impressed GPT-4 with 96.6\% ChatGPT Quality, setting a new SOTA among
open-source LLMs.

\subsection{Siginificance of Phoenix}
\begin{itemize}
    \item We conduct instruction-following adaption in multiple languages, especially for non-Latin languages. To the best of our knowledge, Phoenix is the first open-source multi-lingual democratized ChatGPT, where it uses rich multi-lingual data in the pre-training and instruction-finetuning stages. 
    \item In training Phoenix, we exploited both instruction and conversation data during post-training.\footnote{Koala~\cite{koala_blogpost_2023}, the concurrent work, also used both types of data.} Experimental results demonstrate the effectiveness of using them simultaneously.
    \item Phoenix is among the first-tier Chinese large language models, achieving a performance close to that of OpenAI ChatGPT; its Latin version Chimera is competitive in English. For many other languages, Phoenix is the SOTA open-source large language model. 
    \item  We benchmarked many existing LLMs using automatic and human evaluations. We additionally evaluate the multiple aspects of language generations of LLMs. This is among the first work to evaluate extensive large language models systematically.
\end{itemize}
\section{Overview of existing Democratized ChatGPTs}
\subsection{The tendency to democratize ChatGPT}
\label{sec:related_work}
Since the release of ChatGPT, an increasing number of related models have been developed and published based on the LLaMA~\cite{touvron2023llama} and BLOOM~\cite{workshop2023bloom} models. Other than LLaMA and BLOOM that were \textit{pre-trained} by a massive amount of plain corpora, the recent work tends to focus on \textit{post-training}, which take a pre-trained backbone model (e.g., LLaMA and BLOOM) and skip the first pre-training step. Note that post-training is much computationally cheaper and there affordable to some research teams. 
These post-training based works can be divided into two categories. The first category is instruction-based tuning, and Alpaca~\cite{alpaca} is a notable example. It employs the self-instruction technique~\cite{wang2022selfinstruct} to generate more instructions by the GPT 3.5 model for fine-tuning, resulting in more accurate and contextually relevant outputs. Subsequently, the second category is conversation-based tuning models that utilize the distillation of user interactions with ChatGPT. Vicuna~\cite{vicuna2023} serves as a representative model for this approach, capitalizing on large-scale user-shared dialogue datasets to improve model performance. Aside from a few commercialized, non-open-source models (such as Baidu-Wenxin\footnote{\url{https://yiyan.baidu.com/}}), the majority of popular open-source models follow the principles of these two categories of post-training in their training methodologies and the most representative work are shown in Table~\ref{tab:model_comparison}.

\begin{table}[ht]
\centering
\caption{Comparison of existing popular democratized ChatGPT models.}
\label{tab:model_comparison}
\begin{threeparttable}
\resizebox{1.0\textwidth}{!}{
\begin{tabular}{llc|cc|c|cc|cc|c}
\toprule
 \multirow{2}{*}{Model} &  \multirow{2}{*}{Backbone}  & \multirow{2}{*}{\#Params} &  \multicolumn{2}{c}{Open-source}  & Claimed  & \multicolumn{4}{c}{Post-training}   & Release  \\
 & &   &  &  & language & \multicolumn{2}{c}{instruction}& \multicolumn{2}{c}{conversation}   &  date\\
 & &   & model & data & & data & lang &  data & lang  &\\\midrule
ChatGPT & unknown & unknown & \xmark & \xmark & multi &  &  &&  & 11/30/22\\
Baidu-Wenxin  & unknown& unknown & \xmark & \xmark & zh &    & && & 03/16/23\\  
ChatGLM \footnotemark ~\cite{zeng2023glm-130b} & GLM & 6B  & \cmark\tnote{1} & \xmark& en/zh & && &  & 03/16/23\\
\hline
Alpaca~\cite{alpaca} & LLaMA & 7B & \cmark & \cmark & en & 52K & en & \xmark&\xmark&  03/13/23\\
Dolly  \tnote{2}  & GPT-J & 6B&  \cmark &  \cmark & en & 52K & en &\xmark&\xmark & 03/24/23\\
BELLE~\cite{belle2023exploring}& BLOOMZ & 7B &  \cmark &  \cmark& zh& 1.5M  & zh   &\xmark&\xmark  & 03/26/23\\
Guanaco  & LLaMA& 7B & \cmark  &  \cmark &4\tnote{4} & 534K\tnote{3} & 4\tnote{4}&\xmark&\xmark  & 03/26/23\\ 
Chinese-Alpaca~\cite{chinese-llama-alpaca} & LLaMA & 7/13B & \cmark & \cmark  & en/zh & 2M/3M & en/zh &\xmark&\xmark & 03/28/23\\
LuoTuo~\cite{Luotuo} & LLaMA & 7B &  \cmark &  \cmark& zh & 52k &  zh  &\xmark&\xmark& 03/31/23\\
Vicuna~\cite{vicuna2023} & LLaMA & 7/13B & \cmark & \cmark\tnote{5} & en &\xmark&\xmark &70K & multi\tnote{6}&  03/13/23\\
Koala  & LLaMA & 13B & \cmark & \cmark& en& 355K & en &117K& en & 04/03/23\\
BAIZE~\cite{xu2023baize} & LLaMA& 7/13/30B &  \cmark &  \cmark& en &  52K &  en & 111.5K & en& 04/04/23 \\\hline
\textbf{Phoenix} & BLOOMZ& 7B &  \cmark &  \cmark& multi& 267K & 40+ & 189K& 40+ & 04/08/23 \\
\textbf{Latin Phoenix (Chimera)} & LLaMA & 7B/13B &  \cmark &  \cmark&Latin& 267K & 40+ & 189K& 40+& 04/08/23 \\
\bottomrule
\end{tabular}}
\begin{tablenotes}
      \item [1] Only release the weights.
      \item [2] Dolly 2.0, based on the Pythia-12 b model, was published on 04/12.
      \item [3] 32,880 chat dialogues without system input and 16,087 chat dialogues with system input. 
      \item [4]English, Simplified Chinese, Traditional Chinese (Taiwan, Hong Kong), Japanese, Deutsch.
      \item [5] They only claimed that ShareGPT is the data source but did not provide the files. 
      \item [6] This dataset is collected from ShareGPT, mainly in English. 
     \end{tablenotes}
\end{threeparttable}
\end{table}

\footnotetext{\url{https://github.com/THUDM/ChatGLM-6B}}

\paragraph{Instruction-based Tuning}
Although Alpaca~\cite{alpaca} only released a training set consisting of 52K examples generated using the self-referential instruction method, many variant models have been fine-tuned on Alpaca's instruction dataset, including Dolly\footnote{\url{https://huggingface.co/databricks/dolly-v1-6b}} based on GPT-J~\cite{schick2023toolformer} and LuoTuo~\cite{Luotuo}, which is based on LaMMA and is trained on translated versions of the dataset in Chinese. The BELLE model~\cite{belle2023exploring}, on the other hand, followed the self-instruction process of Alpaca and generated a Chinese dataset of 1.5M samples by using 175 manually constructed Chinese seed instructions. It is an optimized and refined version of the BLOOMZ-7B1-mt model~\cite{workshop2023bloom} and more suitable for Chinese culture and background knowledge due to the Chinese dataset. Chinese-alpaca~\cite{chinese-llama-alpaca} adapts English and translated Chinese Alpaca dataset based on LLaMA to support the bi-lingual environment. Some researchers~\cite{peng2023gpt4llm} attempt to use a stronger teacher model to generate instruction data. Furthermore, Guanaco\footnote{\url{https://guanaco-model.github.io/}} adds external more languages (English Simplified Chinese, Traditional Chinese, Japanese, and Deutsch) entries with Alpaca dataset and is trained based on LLaMa to show the potential in a multilingual environment. 

\paragraph{Conversation-based Tuning}
Inspired by the impressive results achieved by the Vicuna, training models through distilling data from user-shared chatGPT conversations has become a new trend. However, since Vicuna did not publicly release the dataset samples they used from ShareGPT, most subsequent models had to construct similar datasets by themselves. Based on existing open-sourced instruction datasets, Koala\footnote{\url{https://bair.berkeley.edu/blog/2023/04/03/koala/}} utilized 30K conversation examples from ShareGPT with non-English languages removed and also incorporated the English question-answering dataset HC3~\cite{guo2023close}. BAIZE~\cite{xu2023baize} used a novel pipeline that generates a high-quality multi-turn conversation corpus containing 111.5K samples by having ChatGPT engage in a conversation with itself as the training dataset.

\subsection{Multilingual Capabilities of Democratized ChatGPT Models}

%

Currently, most large language models are designed specifically for Latin languages, with English being the primary focus, see many LLaMa based language models in  Table \ref{tab:support}. This limitation hinders their widespread use worldwide, particularly in countries where non-Latin languages are spoken. While some models based on the LLaMA backbone incorporate a small amount of non-Latin data in the \textit{post-training} stage, their multilingual capabilities are primarily derived from the massive \textit{pre-training} corpora, meaning that any models trained from the LLaMA backbone are somehow restricted to Latin and Cyrillic languages. Guanaco is one such example, where Chinese and Japanese are added in the post-training stage, but the LLaMA backbone does not include Chinese and Japanese corpora, and the vocabulary used is not fully supportive of these languages. To this end, Chinese-alpaca~\cite{chinese-llama-alpaca} made a lot of efforts to augment the LLaMA vocubulary with Chinese tokens.

On the other hand, BELLE utilizes Bloom as its backbone, which is more versatile in accommodating a wide range of languages, including non-Latin scripts. However, BELLE is specifically fine-tuned for Chinese, which restricts its multilingual capabilities. In contrast, Phoenix is a Bloom-based large language model that supports multiple languages both during pre-training and post-training. Meanwhile, Chimera is the Latin version of Phoenix that employs a Latin-based backbone (LLaMA). As expected, later experiments have shown that Chimera largely underperforms Phoenix due to the lack of non-Latin corpora in the backbone pre-training.

\begin{table}[h]
\centering
\caption{The main language support of open-source models. \ccmark~denotes the given language is supported by the  \textit{pre-trained} backbone  while \cccmark~is for \textit{post-training} support. Models with \textbf{both marks} (i.e., \ccmark \cccmark) will be considered capable of handling the given language. }
\label{tab:support}
\resizebox{1.0\textwidth}{!}{
\begin{tabular}{lllllll|lllll}
\toprule
Language & \multicolumn{6}{c}{Latin} & \multicolumn{4}{c}{Non-Latin}      \\
& English & French & Spanish  & Portuguese    & Italian   & Deutsch   & Chinese & Arabic   & Japanese  & Korean      \\
\midrule
Dolly & \ccmark \cccmark &   &   &   &  &  &  &  \\
Alpaca & \ccmark \cccmark & \ccmark & \ccmark & \ccmark & \ccmark & \ccmark &  & \\
Koala & \ccmark \cccmark & \ccmark & \ccmark & \ccmark & \ccmark & \ccmark &  & \\
Baize & \ccmark \cccmark & \ccmark  & \ccmark  & \ccmark  & \ccmark & \ccmark &  &  \\
Vicuna & \ccmark \cccmark & \ccmark \cccmark & \ccmark \cccmark & \ccmark \cccmark & \ccmark \cccmark& \ccmark \cccmark&  &\\
LuoTuo & \ccmark  & \ccmark  & \ccmark  & \ccmark  & \ccmark & \ccmark & \myspace \cccmark & &  \\
Chinese-Alpaca & \ccmark \cccmark & \ccmark & \ccmark & \ccmark & \ccmark & \ccmark &  \myspace \cccmark &  \\
Guanaco & \ccmark \cccmark & \ccmark  & \ccmark  & \ccmark & \ccmark & \ccmark \cccmark &  \myspace \cccmark & & \myspace \cccmark \\
BELLE & \ccmark & \ccmark  & \ccmark & \ccmark & \ccmark & \ccmark &  \ccmark \cccmark &  \ccmark & \ccmark & \ccmark \\
\midrule
Phoenix & \ccmark \cccmark & \ccmark \cccmark & \ccmark \cccmark & \ccmark \cccmark & \ccmark \cccmark & \ccmark \cccmark & \ccmark \cccmark & \ccmark \cccmark & \ccmark \cccmark  & \ccmark \cccmark  \\
Latin Phoenix (Chimera) & \ccmark \cccmark & \ccmark \cccmark & \ccmark \cccmark & \ccmark \cccmark & \ccmark \cccmark & \ccmark  \cccmark &  \myspace \cccmark &  \myspace \cccmark & \myspace \cccmark  &  \myspace \cccmark  \\
\bottomrule
\end{tabular}
}
\end{table}

\section{Methodology}

\subsection{Dataset Construction}
We collected our data from two sources: instruction data and user-shared conversations. We followed self-instruction~\cite{wang2022selfinstruct} to construct the instruction data and followed Vicuna~\cite{vicuna2023} to collect the user-share conversation data. To ensure the diversity of instructions and languages, we propose using a role-centric approach to construct instruction data and translate the instruction data to multiple languages. The details of the two types of data are shown as follows:

\subsubsection{Instruction Data}
We use three groups of instruction data as listed below.
\begin{itemize}
    \item \textbf{Collected multi-lingual Instructions}: We used the 52K instructions collected in Alpaca~\cite{alpaca}, where each sample includes \textit{instrction} (the task descriptions for large language models), \textit{input} (the optional context for the instruction task), and \textit{output} (the answers generated by large language models). For the \textit{output}, we used the GPT-4-version ones released by \cite{peng2023gpt4llm}, including both the English (Alpaca-gpt4-en) and Chinese (Alpaca-gpt4-zh) answers.
    \item \textbf{Post-translated multi-lingual instruction}: Based on the data described above, we have collected additional instructions in multiple other languages (such as French, Spanish, Portuguese, and Italian, etc.) through translation. This collection includes two parts: one part (Alpaca-ml-gpt4-post-translation) involves the complete translation of the instructions and outputs generated by GPT-4, while the other part (Alpaca-ml-gpt35-post-output) involves only translating the instructions of GPT-4 and generating the output using the GPT-3.5 model, which helps alleviate the issue of translation difficulties when specific languages are required for output generation. The process is formalized in Algorithm~\ref{alg:post-translation}. We acknowledge that translation might distort instructions, especially when instructions are language-specific. For example, a prompt {\tt write a Chinese Poet, like seven character quatrains} cannot be properly answered by another language. We leave dealing with the translation distortion as future work.
    \item \textbf{Self-generated User-centered multi-lingual instructions}
Besides the above instructions, we also build some instruction data by ourselves (User-centered instructions). The main difference is that our instructions are driven by a given role (user)  set. {\tt role}  could be either the executor or the submitter of a given instructor. It is possible to leave {\tt role} empty to improve robustness. We also translate them into 40 languages according to the population of speakers\footnote{\url{https://en.wikipedia.org/wiki/List_of_languages_by_total_number_of_speakers}}. The process is formalized in Algorithm~\ref{alg:user-centered-instructions}.
\end{itemize}    
\begin{algorithm}[h]
  \SetAlgoLined
  \KwIn{ Instruction Data $\mathbb{D}$, containing many instruction pairs ${(\tt instruction}, {\tt input}) \in \mathbb{D}$  }
  \KwOut{Translated multi-lingual triplets ${(\tt instruction}^\prime, {\tt input}^\prime, {\tt output}^\prime) \in \mathbb{D}^\prime$  }
  \ForEach{instruction pair}{
   Sample another language ${\tt lang}$ based on the general language distribution\;
    Translate ${(\tt instruction}, {\tt input})$ into the sampled language:${(\tt instruction^\prime}, {\tt input^\prime})$\;
    Generate ${\tt output}^\prime$ based on the translated instruction ${(\tt instruction^\prime}, {\tt input^\prime})$\;
  }
\caption{Post-translation for multi-lingual instruction}
\label{alg:post-translation}
\end{algorithm}
\begin{algorithm}[h]
\caption{Generation of user-centered instructions}
\label{algo:chatgpt}
\SetAlgoLined
\KwIn{None}
\KwOut{User-centered instruction quadruples $  \{({\tt role}, {\tt instruction}, {\tt input}, {\tt ouput}$) \} }
\textbf{Step 1:} Build a role set using  a well-design ChatGPT prompt and manual efforts;

\textbf{Step 2:} Manually build some  seed triplets  $\{({\tt role}, {\tt instruction}, {\tt input})\}$ for each {\tt role};

\textbf{Step 3:} Generate more triplets using the seed triplets in an in-context few-shot fashion;

\textbf{Step 4:} \ForEach{instruction triplet}{Predict its {\tt output} based on the  triplet  $({\tt role}, {\tt instruction}, {\tt input})$.}
\label{alg:user-centered-instructions}
\end{algorithm}

\subsubsection{Conversation Data} 
We mainly use ChatGPT-distilled conversation to adapt our language model for chatting. There are two sources of  ChatGPT-distilled conversation data, and we translate them into 40 languages according to the population of speakers.  

\textbf{ShareGPT } ShareGPT\footnote{\url{https://sharegpt.com/}} is a Chrome extension that allows users to conveniently share their ChatGPT conversations. The data could be downloaded from Huggingface Datasets.\footnote{\url{https://huggingface.co/datasets/philschmid/sharegpt-raw}}

\textbf{Discord ChatGPT channel} Discord is a free messaging software and digital platform for communities designed for gamers, educators, friends, and business people to communicate via chat, images, videos, and audio. The ChatGPT channel is the place for users to submit prompts in order to receive responses. ShareGPT is previously used by Vicuna~\cite{vicuna2023} while Discord ChatGPT channel is shared in our project.\footnote{\url{https://github.com/FreedomIntelligence/LLMZoo}} Unlike Koala, we do not exclude non-English conversation data.

\subsection{Dataset Statistics}

\begin{table}[h]
\centering
\caption{The statistics on the components of our dataset.}
\resizebox{1.0\textwidth}{!}{
\begin{tabular}{llrrrr}
\toprule
Type & Dataset & Samples & Turns & Avg. tokens/sample & Avg. tokens/turn \\ \midrule
\multirow{5}{*}{Instruction} & Alpaca-gpt4-en & 52K & 52K & 198.60 & 198.60 \\ 
&Alpaca-gpt4-zh & 4.9K & 4.9K & 338.92 & 338.92 \\ 
&Alpaca-ml-gpt4-post-translation & 51K & 51K & 543.39 & 543.39 \\ 
&Alpaca-ml-gpt35-post-output & 49K & 49K & 435.11 & 435.11 \\ 
&User-centered instructions &  65K & 65K & 474.60 & 474.60 \\ \midrule
\multirow{2}{*}{Conversation} &ShareGPT & 90K & 655K & 3835.30 & 527.06 \\
&Discord & 8K & 18K & 487.68 & 232.75 \\ \midrule
&ALL & 465K & 939K & 982.35 & 486.04 \\ \bottomrule
\end{tabular}}
\label{tab: data_static}
\end{table}

Table~\ref{tab: data_static} provides a comprehensive overview of the statistics for the various sub-datasets within our dataset. For each sub-dataset, we present the number of samples, the number of turns, the average tokens per sample, and the average tokens per turn. The overall statistics, encompassing all sub-datasets, are summarized in the row labeled ``ALL''. This information allows for a clear comparison and understanding of the various components within our dataset, which is crucial for evaluating the performance and characteristics of our models under investigation.

\begin{figure}[h]
    \centering
    \includegraphics[width=0.98\textwidth]{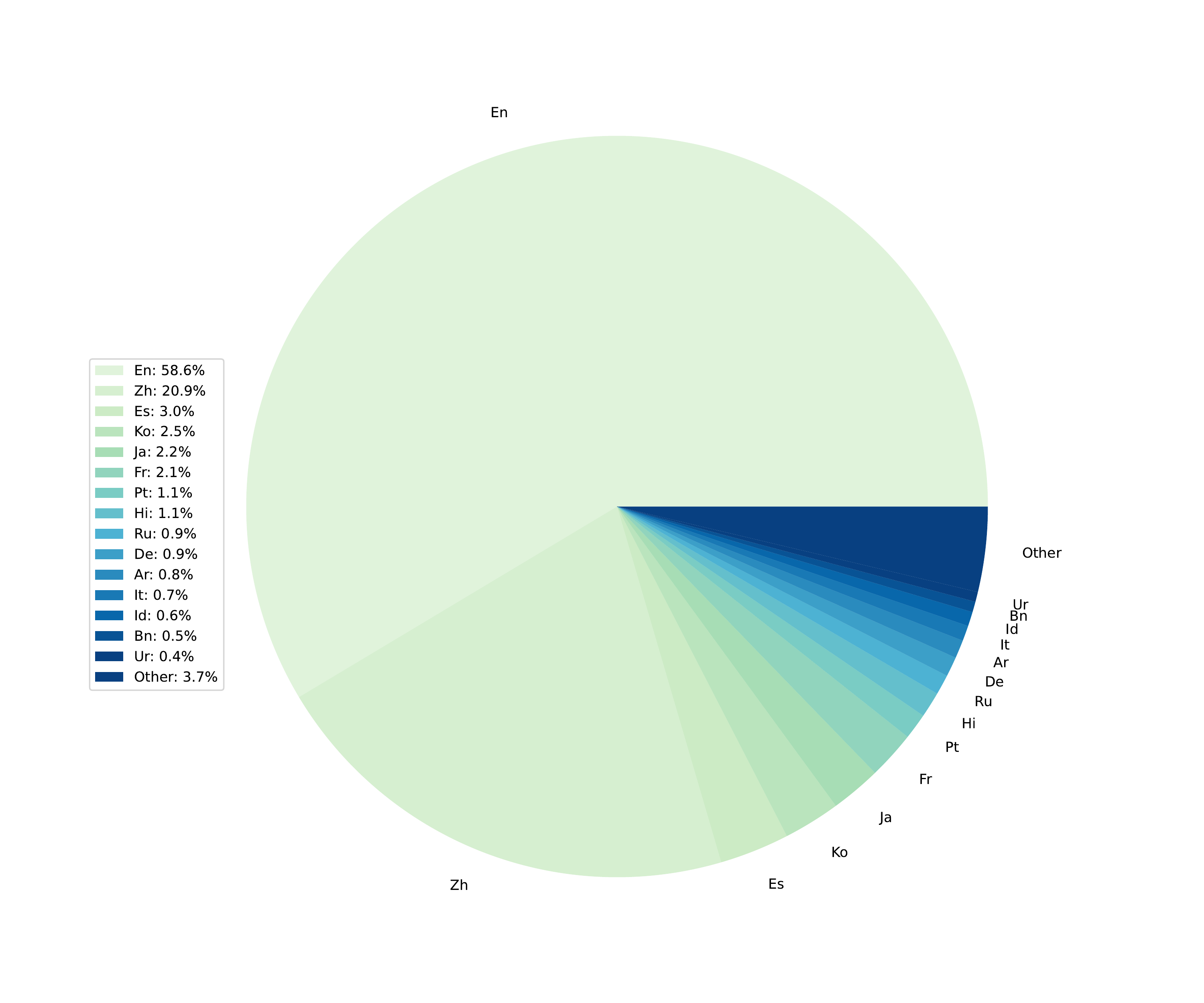}\vspace{-25pt}
    \caption{Language Distribution in Our Dataset: Top 15 Languages Represented out of 133.}
    \label{fig:language_distribution}
\end{figure}

Figure~\ref{fig:language_distribution} provides a visual representation of the language distribution in our dataset, emphasizing the top 15 languages. The short name of languages is from ISO 639-1\footnote{\url{https://en.wikipedia.org/wiki/List_of_ISO_639-1_codes}}. The data reveals that English and Chinese constitute the majority of the dataset, with a combined proportion of approximately 79.5\%. The other 13 languages in the top 15 together make up the remaining 17.8\%, demonstrating a diverse range of languages in the dataset.

\subsection{Training Details}
The models are implemented in PyTorch using the Huggingface Transformers package\footnote{\url{https://github.com/huggingface/transformers}}. We set the max context length to 2,048. We train the model with the AdamW optimizer, where the batch size and the number of epochs are set to 256 and 3, respectively. The learning rate and weight decay are set to 2e-5 and 0, respectively. The model using BLOOMZ as the backbone is called `Phoenix' while that using LLaMA is called `Chimera'.
\section{Evaluation}
\subsection{Challenges}
Assessing the performance of AI chatbots is a challenging task that requires a comprehensive evaluation of language coherence, comprehension, reasoning ability, and contextual awareness. Although \cite{liang2022helm} has elaborated an exhaustive study on evaluating LLMs on existing benchmarks, it may no longer be adequate. We summarize the existing \textbf{evaluation dilemma} for LLMs with the following three-fold challenges:
\begin{itemize}
    \item \textbf{Not-blind}: Test data or similar data in benchmark might be seen by LLMs during pre-training of supervised fine-tuning.
    \item \textbf{Not-static}: The ground truth is not static, e.g., {\tt tell a joke about Donald Trump}.
    \item  \textbf{Incomplete testing path coverage}: Unlike path coverage of codes in Software Engineering, full coverage of testing cases is impossible since user prompts are multi-faced. 
\end{itemize}

To address these challenges, we present an evaluation framework based on GPT-4/GPT-3.5 Turbo API to automate chatbot performance assessment.

\subsection{Evaluation Protocol}
\paragraph{Baselines}
To validate the performance of Phoenix, we first compare it with existing instruction-tuned large language models in Chinese and English, including GPT-3.5 Turbo, ChatGLM-6b, Wenxin, BELLE-7b-2m, Chinese-Alpaca 7b/13b, Vicuna-7b/13b\footnote{We used the latest version of Vicuna models released in 04/13/2023.}.
Besides, we evaluate our models on more Latin (e.g., French, Spanish, and Portuguese) and non-Latin languages (e.g., Arabic, Japanese, and Korean) to show the multi-lingual ability, where we mainly compare our models with GPT-3.5 Turbo and a multi-lingual instruction-tuned model, Guanaco.

\paragraph{Metrics}
Following the evaluation of the Vicuna~\cite{vicuna2023}, we assess our model by testing it on a set of 80 questions spanning 8 distinct categories. Additionally, we include two more categories, namely reasoning, and grammar, bringing the total number of questions to 100, spread across 10 categories. 

In order to make our comparison clearer, we conduct a pairwise comparison of the models' absolute performance by strictly following the evaluation settings of Vicuna~\cite{vicuna2023}, such as prompts, decoding hyperparameters, etc. To achieve this, we request GPT-4 to rate the potential answers based on their helpfulness, relevance, accuracy, and level of detail on the 80 English questions in Vicuna's test set. It is important to note that this experiment only contributes to Figure~\ref{fig:comparison}. For the rest of our evaluation, we compare models on our curated 100 questions in different languages.

Limited by the quota of the OpenAI account, we only utilized GPT-4 API as a reviewer to provide an absolute performance score on the test sets in Chinese and English. For the rest of our experiments, we resort to GPT-3.5 Turbo  API. While GPT-3.5 Turbo  can give a proper review of the candidate answer pairs, it is generally less reliable than GPT-4 API. We posit that, aside from potential limitations in the model capacity of ChatGPT itself, assigning scores to two answers with minimal differences in quality would be a challenging task. In addition, we found that GPT-3.5-Turbo tends to assign high scores to both of the two answers, which can lead to the overestimation of model performance. Therefore, we opted for a secondary approach, in which GPT-3.5 Turbo  is tasked with determining which of two answers is better and providing the justification for its choice.
The evaluation prompt we are using for absolute performance and for beat rate are listed below:

\textbf{Performance Ratio:} \texttt{We would like to request your feedback on the performance of two AI assistants in response to the user question displayed above.\textbackslash n Please rate the helpfulness, relevance, accuracy, and level of detail of their responses. Each assistant receives an overall score on a scale of 1 to 10, where a higher score indicates better overall performance.\textbackslash n Please first output a single line containing only two values indicating the scores for Assistant 1 and 2, respectively. The two scores are separated by a space.\textbackslash n In the subsequent line, please provide a comprehensive explanation of your evaluation, avoiding any potential bias and ensuring that the order in which the responses were presented does not affect your judgment.}

\textbf{Beat Rate:} \texttt{We would like to request your feedback on the performance of two AI assistants in response to the user question displayed above.\textbackslash n Please evaluate the given four aspects: helpfulness, relevance, accuracy, level of details of their responses.\textbackslash n Please first clarify how each response achieves each aspect respectively.\textbackslash n Then, provide a comparison of the overall performance between Assistant 1 and Assistant 2, and you need to clarify which one is better than or equal to another. Avoid any potential bias and ensure that the order in which the responses were presented does not affect your judgment.\textbackslash n In the last line, order the two assistants. Please output a single line ordering Assistant 1 and Assistant 2, where `>' means `is better than' and `=' means `is equal to'. The order should be consistent with your comparison. If there is no comparison that one is better, it is assumed they have equivalent overall performance (`=').}

Then we calculate the performance ratio by averaging the scores obtained by each model across our 100 questions while calculating the beat rate by using the number of times that the model wins divided by the sum of the number of times that the model wins and the number of times that the model loses.\footnote{Currently we don't count the cases where the GPT-3.5 Turbo  gives a result of a tie.}
Please refer to Appendix~\ref{app:eval_detail} for further details.

\subsection{Experimental Results}
We first conducted monolingual tests in both English and Chinese. We request GPT-4 to assign a quantitative score to each response on a scale of 1 to 10. Then we calculate the final score for each comparison pair (baseline, Phoenix) by averaging the scores obtained by each model across our 100 questions in the English and Chinese subsets.

\begin{table}[h]
\label{tab:phoenix-gpt4}
\centering
\caption{Benchmarking Phoenix in English and Chinese.  The winner in each competition is in \textbf{bold}. The performance ratio is scored by GPT-4 API, and the beat rate is calculated using GPT-3.5 Turbo.}
\resizebox{1.0\textwidth}{!}{
\begin{tabular}{@{}lrrrr@{}}
\toprule
\multirow{2}{*}{ Comparision}   & \multicolumn{2}{c}{Zh}    & \multicolumn{2}{c}{En}    \\
                         &  Performance Ratio  & Beat Rate  & Performance Ratio  & Beat Rate  \\ \midrule
Phoenix vs. Phoenix (anchor)                & 100      & 50        & 100     & 50        \\ 
\midrule
Phoenix vs. \textbf{GPT-3.5 Turbo}                  & 85.20      & 35.75        & 87.13      & 43.75        \\
Phoenix vs. \textbf{ChatGLM-6b}               & 94.60      & 36.00        & 121.11     & 54.50        \\
Phoenix vs.  \textbf{Baidu-Wenxin}                & 96.80      & 44.00        & -          & -            \\
\textbf{Phoenix} vs.  BELLE-7b-2m              & 122.70     & 65.25        & -          & -            \\
\textbf{Phoenix} vs. Chinese-Alpaca-7b  & 135.30     & 75.75        & -          & -            \\
\textbf{Phoenix} vs. Chinese-Alpaca-13b & 125.20     & 74.50        & -          & -            \\
\textbf{Phoenix} vs. Vicuna-7b                & -          & -            & 121.2         & 53.00        \\
\textbf{Phoenix} vs. Vicuna-13b               & -          & -            & 90.92          & 46.00        \\ \bottomrule
\end{tabular}}
\label{tab:zh-en}
\end{table}

\paragraph{Chinese}
We compared our model with the mainstream Chinese models, as shown in   Table~\ref{tab:zh-en}. It slightly underperforms Baidu-Wenxin and ChatGLM-6b, which are not fully open-source; ChatGLM-6b only provides model weights without training data and details.
Phoenix underperforms ChatGLM-6B, which may be attributed to the fact that we did not conduct reinforcement learning from human feedback (RLHF) like ChatGLM-6B. However, Phoenix achieves comparable performance with Baidu-Wenxin, a commercial and closed-source language model designed solely for Chinese. Given that Baidu-Wenxin may have a larger model and is exclusive to the Chinese, this is a significant achievement for an open-source, democratized ChatGPT developed by academic institutions. It should be noted that neither ChatGLM-6B nor Baidu-Wenxin significantly outperforms Phoenix, as evidenced by our statistical testing.

Phoenix significantly surpasses other popular open-source Chinese models, achieving a performance of over 120\% than them. Specifically, Phoenix achieved 122.70\% of BELLE-7B-2m and 135.30\% of Chinese Alpaca-7b. Notably, even with 7b parameters, Phoenix can outperform the Chinese Alpaca-13b model, achieving a performance level of 125.20\%. It demonstrates that although Phoenix is a multi-lingual LLM, it achieves SOTA performance among all open-source Chinese LLMs.

\begin{figure}[h]
    \centering
    \includegraphics[width=0.9\textwidth]{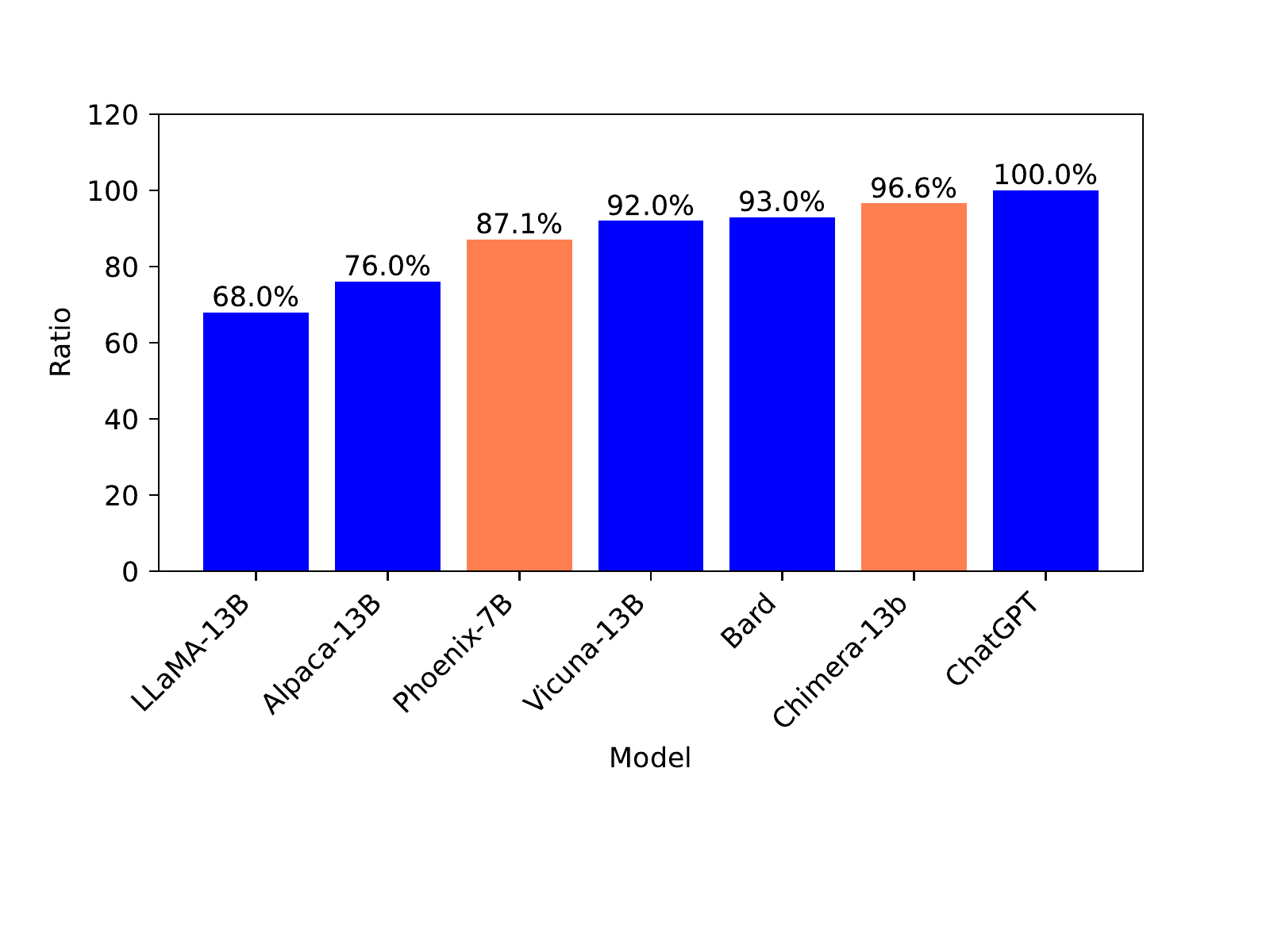}\vspace{-40pt}
    \caption{Relative Response Quality Assessed by GPT-4. }  
    \label{fig:comparison}
\end{figure}

\paragraph{English}
We also compare  Phoenix with Vicuna, ChatGPT, and ChatGLM-6B, which are claimed to work in English. The two columns on the right side of Table~\ref{tab:zh-en} demonstrate the impressive performance in English of our Phoenix. Our model outperforms Vicuna-7b by 21.2\% and ChatGLM-6b by 21.1\%. It is important to note that Phoenix is a multi-lingual LLM. Therefore, compared to Vicuna-13b and ChatGPT, our model still lags behind them in terms of absolute performance in English.
Interestingly, Chimera, as a tax-free Phoenix,  has impressed GPT-4 with 96.6\% ChatGPT Quality, setting a new SOTA in open-source LLMs, see Figure~\ref{fig:comparison}. Note that the evaluation is not rigorous enough. We will conduct human evaluations in the revision.

\paragraph{Other Languages} Table~\ref{tab:multi-lingual} shows the beat rate of Phoenix in multiple languages. In most languages, Phoenix has an absolute advantage over the Guanaco model, which is also multilingual. Due to fewer multilingual taxes, the Latin version of Phoenix (Chimera) performs even better in Latin languages, comparable to the GPT-3.5 Turbo and even slight better than the GPT-3.5 Turbo in French and Spanish, as shown in Table~\ref{tab:multi-lingual-chimera}.

\begin{table}[h]
\centering
\caption{Beat Rate of Phoenix in Multiple Languages.}
\resizebox{1.0\textwidth}{!}{
\begin{tabular}{@{}lcccccccccc@{}}
\toprule
\multirow{2}{*}{Language} & \multicolumn{5}{c}{Latin} & \multicolumn{3}{c}{Non-Latin}      \\
& French & Spanish  & Portuguese    & Italian   & Deutsch    & Arabic   & Japanese    & Korean      \\ \midrule
Phoenix vs. Phoenix (Anchor) & \multicolumn{8}{c}{50}  \\ \midrule
Phoenix vs. GPT-3.5 Turbo  & 41.75  & 34.00   & 32.75  & 19.00   & 10.50   & 30.25  & 25.50   & 7.75  \\
Phoenix vs. Guanaco        & 92.80  & 93.60    & 95.50   & 75.80    & 47.00    & 97.00  & 86.25        & 93.75 \\ \bottomrule
\end{tabular}}
\label{tab:multi-lingual}
\end{table}

\begin{table}[h]
\centering
\caption{Beat Rate of Chimera in Multiple Latin Languages.}
\begin{tabular}{@{}lccccc@{}}
\toprule
Model                         & French    & Spanish    & Portuguese    & Italian    & Deutsch    \\ \midrule
Chimera-13b vs. Chimera-13b (anchor) & \multicolumn{5}{c}{50}                        \\\midrule
Chimera-13b vs. GPT-3.5 Turbo & 54.12 & 52.71 & 40.67 & 47.67 & 45.25 \\
Chimera-13b vs. Guanaco       & 96.00 & 95.50 & 87.50 & 84.80 & 93.00 \\ \bottomrule
\end{tabular}
\label{tab:multi-lingual-chimera}
\end{table}

\subsection{Ablation study }
In order to investigate the contributions of instruction-based and conversation-based data in post-training, we conducted an ablation study. Table~\ref{tab:ablation} shows that adding instruction data is beneficial to Chat-adapted LLMs; instructions achieve 5\%-6\%  relative improvement.

\begin{table}[h]
\centering
\caption{Ablation Study on the Instruction Data.}
\begin{tabular}{@{}lll@{}}
\toprule
                & Phoenix vs. GPT-3.5 Turbo (Zh) & Chimera vs. GPT-3.5 Turbo (En) \\ \midrule
Conversations   & 34.00                          & 39.75                          \\
~ + Instruction & 35.75 {\color{red} $\uparrow 5.1\%$}                         & 42.25  {\color{red} $\uparrow 6.3\%$}                        \\ \bottomrule
\end{tabular}
\label{tab:ablation}
\end{table}

\subsection{Human Evaluation}
To comprehensively evaluate our models, we introduce human evaluation. Specifically, for the 100 questions used for GPT-4, we invite volunteers to rank the results generated by two models for the same question. The results may indicate that one model is better than the other (win and lose), or that they are equally good (tie). We then collect the rankings for all 100 questions of each two model and use them to assess the performance of each model. It is important to note that the two models being evaluated are completely randomized and anonymized, meaning that volunteers are not aware of which model generated each answer. This manual evaluation process ensures that our models are being assessed in a fair and unbiased manner. 

Table~\ref{tab:Human evaluations} presents five different comparisons between Phoenix and other popular models, each consisting of three metrics: win, tie, and lose. Phoenix performs significantly better than open-source Chinese language models (BELLE-7b-2m and Chinese-Alpaca-13b), with absolute advantages of over 50\% (win), respectively. In addition, Phoenix also demonstrates competitive performance with non-open source models (such as ChatGPT and Baidu-Wenxin).
By including human evaluation, we can obtain a more nuanced understanding of our models' performance beyond what quantitative metrics can provide. More examples can be seen in Appendix~\ref{app:eval_detail}.

\begin{table}[h]
\centering
\caption{Human evaluation of the Chinese answers of different models.}
\begin{tabular}{lccc}
\toprule
Comparison & Win & Tie & Lose \\
\midrule
Phoenix vs. \textbf{ChatGPT} & 12 & 35 & 53 \\
\midrule
Phoenix vs. \textbf{Baidu-Wenxin} & 29 & 25 & 46 \\
\midrule
Phoenix vs. \textbf{ChatGLM-6b} & 36 & 11 & 53 \\
\midrule
\textbf{Phoenix} vs. BELLE-7b-2m & 55 & 31 & 14 \\
\midrule
\textbf{Phoenix} vs. Chinese-Alpaca-13b & 56 & 31 & 13 \\
\bottomrule
\end{tabular}
\label{tab:Human evaluations}
\end{table}
\section{Conclusion}
Among the ChatGPT democratization, this work extends LLM to multiple languages. The training philosophy is to combine instruction data and conversation data to tame models to follow instructions in a chat fashion. The resulting multilingual LLM `Phoenix' achieves the SOTA on fully open-source Chinese LLMs. In non-Latin languages, Phoenix outperforms existing open-source LLMs, including Vicuna-13b and Guanaco. Notably, our Latin-version of Phoenix, called `Chimera', impresses GPT-4 with 96.6\% ChatGPT Quality, setting a new SOTA in open-source LLMs.
We believe the proposed models could largely benefit people who could not legally use ChatGPT or related tools, therefore making AI open and equal again.

\section*{Limitations}
Our goal in releasing our models is to assist our community in better replicating ChatGPT/GPT4. We are not targeting competition with other competitors, as benchmarking models is a challenging task. Our models face similar models to those of ChatGPT/GPT4, which include: 1) Lack of common sense; 2) Limited knowledge domain; 3) Biases; 4) Inability to understand emotions; and 5) Misunderstandings due to context. More importantly, the used evaluation in this work is not rigorous  enough. Therefore, we will add more automatic and human evaluations in the future. We only make our models accessible inside our university and SRIBD, see \url{http://10.26.1.135:7860/}.

\newpage
\bibliographystyle{plain}
\bibliography{main}

\clearpage
\appendix

\section{ChatGPT is inaccessible to some countries}
\label{sec:countries}

\begin{figure}
    \centering
    \includegraphics[width=1.0\textwidth]{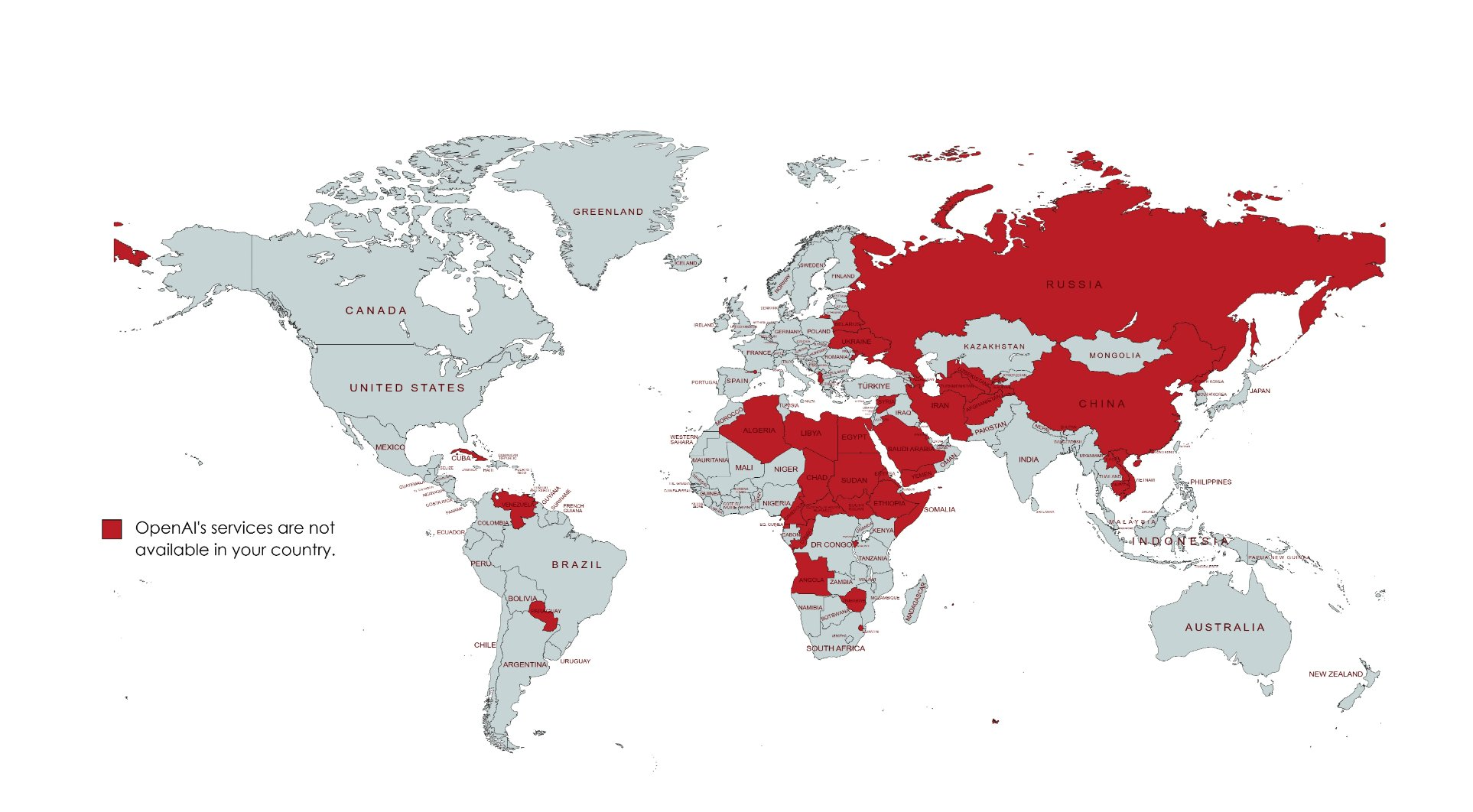}
    \caption{Being Red indicates that in the area ChatGPT is unavailable. Source of the picture: \url{https://i.imgur.com/2fF3Xlh.png}}
    \label{fig:countries}
\end{figure}

As seen in Fig.~\ref{fig:countries}, ChatGPT is unavailable to many countries, most of which are developing countries.

\section{Evaluation Details}
\label{app:eval_detail}

Firstly, we show the prompts and instructions used in our experiments.

\paragraph{GPT-4 review show cases}
To enhance transparency and facilitate comprehension of the reviewing and ranking process, we will present a concrete example below. The question is the 14th question of our curated 100 Chinese questions set. The two answers are from Phoenix-7b and Belle-7b respectively.

\textbf{Input for GPT-4 review:}\\ 

[Question]\\
语言和文化障碍如何影响人们在多元文化社会中的沟通和关系形成？\\

[Assistant 1]\\
语言和文化障碍是人们在多元文化社会中的沟通和关系形成中面临的主要挑战之一。这些障碍可能导致误解、冲突和隔阂，影响人们之间的有效沟通和建立良好关系。
语言障碍可能会导致人们难以理解彼此的语言和表达方式，从而影响交流的流畅性。此外，不同文化背景的人们可能使用不同的词汇、表达方式和习惯用语，这些差异可能导致理解上的障碍。例如，如果两个人来自不同的文化背景，他们可能不太了解对方的幽默感或社交礼仪，这可能会导致误解和冲突。
文化障碍是指人们在跨文化交流中遇到的文化差异，这些差异可能会影响人们对彼此的看法和反应。例如，一项研究发现，西方国家的人可能更倾向于使用个人化的交流方式，而来自亚洲国家的受访者可能更倾向于使用正式和尊重的交流方式。这些差异可能导致冲突和误解，尤其是在处理敏感话题时。
此外，人们在多元文化社会中可能面临不同的价值观和信仰，这可能会导致不同的信仰和行为方式的差异。这些差异可能导致对彼此的不信任和冲突，影响人们之间的关系。
总之，语言和文化障碍可能会导致人们在多元文化社会中难以沟通和建立良好的关系。为了克服这些障碍，人们需要尊重彼此的差异，学习如何在不同文化背景下进行有效的沟通。此外，跨文化交流和理解也是促进多元文化社会和谐发展的重要因素之一。\\

[End of Assistant 1]\\

[Assistant 2]\\
在多元文化社会中，人们面临的语言和文化障碍可能会影响他们的沟通和关系形成。以下是一些可能的影响：

1. 理解不同文化之间的差异：语言和文化障碍可能会导致人们难以理解不同文化之间的差异和不同表达方式。这可能导致误解、冒犯或冲突。

2. 建立和维护关系：在多元文化社会中，人们可能需要努力理解不同文化之间的价值观和信仰。这可能会导致人们在社交场合中感到不舒服或困惑，从而影响他们建立和维护关系。

3. 决策和行动：在多元文化社会中，人们可能需要权衡不同文化之间的利益和风险。由于语言和文化障碍，人们可能无法准确地了解这些利益和风险，从而做出决策或采取行动。

4. 归属感和身份认同：在多元文化社会中，人们可能会感到不确定或不归属感，因为他们无法了解自己的身份认同。语言和文化障碍可能会导致人们难以表达自己的感受和观点，从而影响他们的身份认同。

语言和文化障碍可能会影响人们在多元文化社会中的沟通和关系形成，需要人们采取措施来克服这些障碍，以更好地适应和融入不同的文化环境中。\\

[End of Assistant 2]\\

[System]\\
We would like to request your feedback on the performance of two AI assistants in response to the user question displayed above.

Please rate the helpfulness, relevance, accuracy, and level of detail of their responses. Each assistant receives an overall score on a scale of 1 to 10, where a higher score indicates better overall performance.

Please first output a single line containing only two values indicating the scores for Assistant 1 and 2, respectively. The two scores are separated by a space.

In the subsequent line, please provide a comprehensive explanation of your evaluation, avoiding any potential bias and ensuring that the order in which the responses were presented does not affect your judgment.\\

\textbf{GPT-4 response:}\\

8 7 \\
Assistant 1的回答更详细和全面，提供了语言和文化障碍对沟通和关系形成的影响的具体例子，如幽默感、社交礼仪、交流方式等。此外，Assistant 1还强调了克服这些障碍的重要性，并提出了尊重差异和学习有效沟通的建议。而Assistant 2的回答虽然也涵盖了一些相关影响，但没有提供具体的例子，使得回答相对较为抽象。因此，Assistant 1的回答在帮助度、相关性、准确性和细节方面表现得更好。\\

\paragraph{GPT-3.5-turbo review show cases}
The difference to gpt-4 review mainly lies in the prompt, which corresponds to the content after [system].

\textbf{Input for GPT-3.5 review:}\\ 

[Question]\\
How do language and cultural barriers affect people's communication and relationship formation in a multicultural society?\\

[Assistant 1]\\
Language and cultural barriers can affect people's communication and relationship formation in a multicultural society in several ways:
1. Communication barriers: When people from different cultures speak different languages, they may have difficulty communicating effectively with each other. This can lead to misunderstandings, conflicts, and even social isolation.
2. Conflicts over differences: When people from different cultures have different values, beliefs, and customs, they may find it difficult to accept or respect each other's differences. This can lead to misunderstandings, conflicts, and even animosity.
3. Difficulty forming relationships: When people from different cultures have difficulty communicating or find it difficult to accept each other's differences, they may have difficulty forming meaningful relationships.
4. Isolation: In a multicultural society, people may feel isolated from their peers who share similar cultural backgrounds. This can lead to feelings of loneliness, disconnection, and even depression.
5. Ethnic group dynamics: In multicultural societies, ethnic groups may have their own distinct cultures and languages. This can lead to conflicts over power and resources, and even violence.
Overall, language and cultural barriers can create challenges for people in a multicultural society. However, with effective communication and cultural understanding, people can overcome these challenges and build strong, positive relationships with one another.\\

[End of Assistant 1]\\

[Assistant 2]\\
Language and cultural barriers can significantly affect people's communication and relationship formation in a multicultural society. When people don't share a common language, it becomes challenging to understand and express their thoughts and feelings, which could lead to misunderstandings, conflicts, and break down in communication.

Similarly, people from different cultures may have different values, beliefs, customs, and behaviors that could influence how they perceive and interact with others. These differences could cause a lack of understanding, prejudice, stereotyping, and discrimination, making it challenging to form relationships based on mutual respect and appreciation.

However, when people make an effort to understand and accept cultural differences, they can build stronger bonds and relationships, leading to improved communication and harmonious coexistence. It's crucial to promote language and cultural competence in a multicultural society to enhance effective communication and promote positive relationship formation.\\

[End of Assistant 2]\\

[System]\\
We would like to request your feedback on the performance of two AI assistants in response to the user question displayed above.

Please evaluate the given four aspects: helpfulness, relevance, accuracy, level of details of their responses.

Please first clarify how each response achieves each aspect respectively.

Then, provide a comparison on the overall performance between Assistant 1 and Assistant 2, and you need to clarify which one is better than or equal to another. Avoid any potential bias and ensuring that the order in which the responses were presented does not affect your judgment.

In the last line, order the two assistants. Please output a single line ordering Assistant 1 and Assistant 2, where '>' means 'is better than' and '=' means 'is equal to'. The order should be consistent to your comparison. If there is not comparision that one is better, it is assumed they have equivalent overall performance ('=').\\

\textbf{GPT-3.5 response:}\\

Assistant 1:

- Helpfulness: The response is helpful as it provides a clear explanation of how language and cultural barriers can affect communication and relationship formation in a multicultural society.

- Relevance: The response is relevant as it directly addresses the question asked.

- Accuracy: The response is accurate as it highlights the potential challenges that language and cultural barriers can create in a multicultural society.

- Level of details: The response provides a good level of detail, covering several ways in which language and cultural barriers can affect communication and relationship formation.
\\

Assistant 2:

- Helpfulness: The response is helpful as it provides a clear explanation of how language and cultural barriers can affect communication and relationship formation in a multicultural society.

- Relevance: The response is relevant as it directly addresses the question asked.

- Accuracy: The response is accurate as it highlights the potential challenges that language and cultural barriers can create in a multicultural society.

- Level of details: The response provides a good level of detail, covering the challenges that language and cultural barriers can create and the importance of promoting language and cultural competence.

Overall, both Assistant 1 and Assistant 2 provide helpful, relevant, accurate, and detailed responses to the question. However, Assistant 2 provides a more comprehensive explanation of the challenges and solutions related to language and cultural barriers in a multicultural society. Therefore, the order is: Assistant 2 > Assistant 1.

\end{CJK}
\end{document}